\documentclass[10pt,twocolumn,letterpaper]{article}

\usepackage{wacv}
\usepackage{times}
\usepackage{epsfig}
\usepackage{graphicx}
\usepackage{amsmath}
\usepackage{amssymb}

\usepackage{multirow}
\usepackage{subfigure}
\usepackage{caption}
\usepackage{cite}
\usepackage[accsupp]{axessibility}

\def\wacvPaperID{0420} % Enter the WACV Paper ID here

\wacvfinalcopy % *** Uncomment this line for the final submission

\ifwacvfinal
\fi

\ifwacvfinal
\usepackage[breaklinks=true,colorlinks,bookmarks=false]{hyperref}
\else
\usepackage[pagebackref=true,breaklinks=true,colorlinks,bookmarks=false]{hyperref}
\fi

\begin{document}

%%%%%%%%% TITLE
\title{SEGA: Semantic Guided Attention on Visual Prototype for Few-Shot Learning}

% \author{Fengyuan Yang\\
% \and

\author{Fengyuan Yang$^{1,2}$, Ruiping Wang$^{1,2,3}$, Xilin Chen$^{1,2}$\\
	$^{1}$Key Laboratory of Intelligent Information Processing of Chinese Academy of Sciences (CAS),\\
	Institute of Computing Technology, CAS, Beijing, 100190, China\\
	$^{2}$University of Chinese Academy of Sciences, Beijing, 100049, China\\
	$^{3}$Beijing Academy of Artificial Intelligence, Beijing, 100084, China\\
	{\tt\small fengyuan.yang@vipl.ict.ac.cn, \{wangruiping, xlchen\}@ict.ac.cn}
	%\and
}

\maketitle

\ifwacvfinal
\thispagestyle{empty}
\fi

%%%%%%%%% ABSTRACT
\begin{abstract}
   Teaching machines to recognize a new category based on few training samples especially only one remains challenging owing to the incomprehensive understanding of the novel category caused by the lack of data. However, human can learn new classes quickly even given few samples since human can tell what discriminative features should be focused on about each category based on both the visual and semantic prior knowledge. To better utilize those prior knowledge, we propose the \textbf{SEmantic Guided Attention (SEGA)} mechanism where the semantic knowledge is used to guide the visual perception in a top-down manner about what visual features should be paid attention to when distinguishing a category from the others. As a result, the embedding of the novel class even with few samples can be more discriminative. Concretely, a feature extractor is trained to embed few images of each novel class into a visual prototype with the help of transferring visual prior knowledge from base classes. Then we learn a network that maps semantic knowledge to category-specific attention vectors which will be used to perform feature selection to enhance the visual prototypes. Extensive experiments on miniImageNet, tieredImageNet, CIFAR-FS, and CUB indicate that our semantic guided attention realizes anticipated function and outperforms state-of-the-art results.
\end{abstract}
\vspace{-0.45cm}
%%%%%%%%% BODY TEXT
\section{Introduction}

Object recognition has been significantly improved in the past decade with the rapid growth of data scales and the help of deep learning methods \cite{krizhevsky2017imagenet,simonyan2014very,szegedy2015going}. However, the frequency distribution of visual categories generally presents the form of long-tailed distribution \cite{reed2001pareto} which means it is difficult to collect a sufficient number of samples for categories on the tail part. Even for categories on the head part, a large scale of image annotation is also a heavy and expensive job. Therefore, we study this kind of more realistic task named few-shot learning (FSL) which means learning new categories based on few labeled samples.

\begin{figure}[t]
	\begin{center}
		\includegraphics[width=1.0\linewidth]{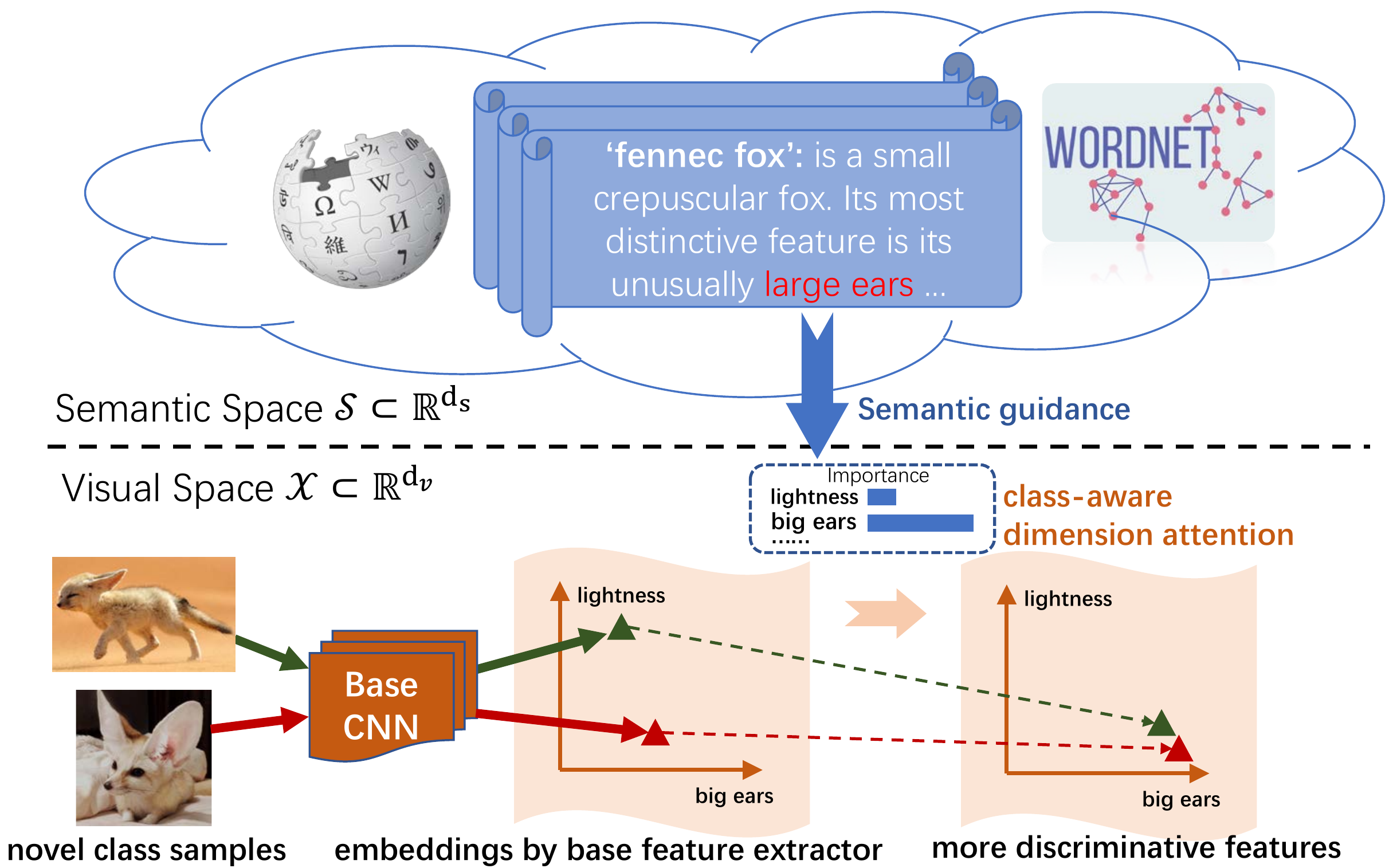}
	\end{center}
	\vspace{-0.45cm}
	\caption{The illustration diagram shows the motivation of ours semantic guided attention. In few-shot learning, the visual cognition of novel class is incomprehensive given few labeled images. However, semantic knowledge can give guidance about what key feature dimensions of this category should be focused on. By applying semantic attention to visual features, we can have more discriminative recognition of the novel class.}
	\label{fig:overview}
	\vspace{-0.55cm}
\end{figure}

\begin{figure*}[t]
	\begin{center}
		%\fbox{\rule{0pt}{2in} \rule{.9\linewidth}{0pt}}
		\includegraphics[width=1.0\linewidth]{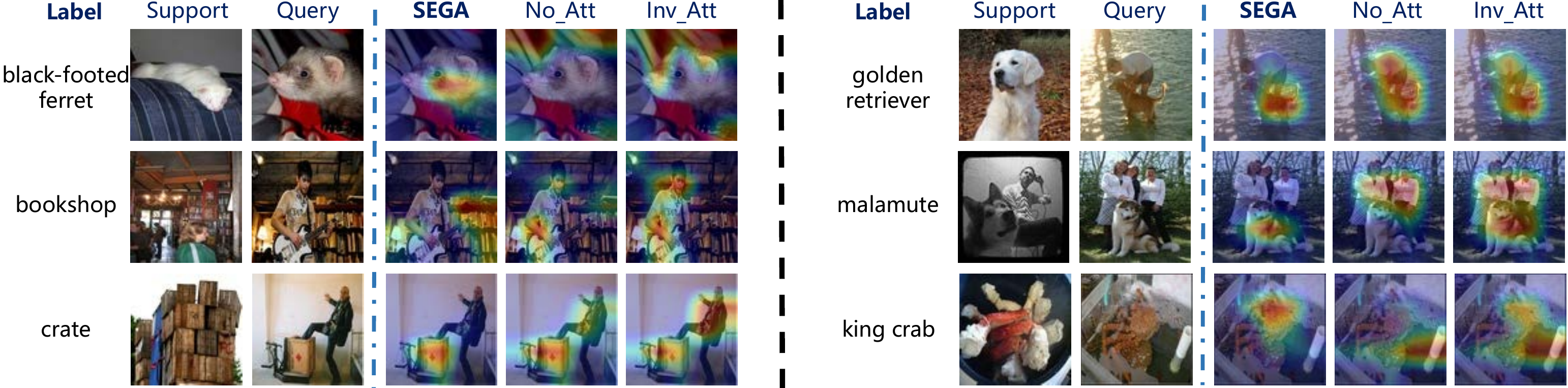}
	\end{center}
	\vspace{-0.45cm}
	\caption{The Grad-CAM visualization testing on miniImageNet's unseen classes under 1-Shot scenario. Column ``\emph{Support}" gives the only training sample for each novel class. Column ``\emph{Query}" shows the query image of this novel class. The next three columns show the query image's Grad-CAM visualization when applying our semantic guided attention(``\emph{SEGA}"), not applying any attention(``\emph{No\_Att}") and applying the inversed version of our semantic guided attention(``\emph{Inv\_Att}") respectively. Warmer color with a higher value.}
	\label{fig:gradcam}
	\vspace{-0.6cm}
\end{figure*}

One of the main challenges of few-shot learning is that when only given few labeled samples, the recognizer cannot get the comprehensive recognition of the novel class. To deal with this challenge, prior knowledge is of vital importance since the reason why we human beings can learn new categories quickly and efficiently is that we have already learned so many base categories before. Therefore in few-shot learning, we usually transfer knowledge from base classes with a large number of labeled images to the target novel classes with a small number of labeled images. Thus most previous works have focused on how to transfer visual prior knowledge efficiently from base classes to novel classes \cite{finn2017model,gidaris2018dynamic,snell2017prototypical,vinyals2016matching}. However, the semantic prior knowledge plays a pivotal role in human learning too. For example, the category name along with its semantic explanation is always mentioned when parents teach children to recognize the animals in the atlas, from which children can have more comprehensive recognition about the new category and the relation between this category and categories learned before.  More specifically, from the semantic guidance human can get what key features of the category should be focused on and what noisy features should be ignored as illustrated in Figure \ref{fig:overview}, which could help us to tackle the above challenge of few-shot learning.

%Utilizing its semantic information can give more knowledge about the category itself which is essential for the category learning. This kind of idea that using semantic information to facilitate visual recognition task comes from a related research area named zero-shot learning \cite{farhadi2009describing,lampert2013attribute,lampert2009learning}.  Most recently, there emerge works which use semantic knowledge in few-shot learning \cite{xing2019adaptive,chen2019multi,schwartz2019baby}. However, which semantic knowledge to utilize and how to utilize them in a more reasonable and efficient way has still been a largely under explored domain and is worth further discussing.

Our motivation for using semantic guidance is derived from cognitive neurosciences. The reason why a human can perform \textit{``object constancy"}, which is the ability to identify objects across changes in the detailed context including illumination, object pose, and background \cite{ullman1996high}, is that our object recognition system can tell the key discriminative features concerning each category. For the categories with enough samples, the key features can be concluded from a large number of images. But as for the categories with few images, human can also get which key features should be focused on directly from the high-level guidance of semantic information such as the class name. Apart from cognitive neurosciences, we can also find a similar idea in contrastive learning \cite{he2020momentum,oord2018representation} which suggests that there is no need to precisely reconstruct the dollar bill but only need to tell key features of the bill to distinguish it from other objects \cite{epstein2016empty}. We cannot rely only on the semantic information, which is compact, to directly reconstruct the visual prototype well enough. However, it makes sense to use the semantic information to get which key features should be paid attention to when performing classification.

Therefore, we propose a more human-like way to utilize label semantic knowledge inspired by the above theories. A feature extractor is trained to transfer the visual prior knowledge from base classes to obtain a feature space. After that, our framework learns a network that maps the class semantic knowledge to the class-specific attention which implies key dimensions of visual features. Then during testing, the visual feature attention of novel classes can be obtained by taking the names of novel classes as input to the above mapping. Finally, the attention will be applied to the visual prototype to highlight the key features so that a more discriminative representation for the novel class can be obtained.

More specifically, we take the first Grad-CAM \cite{selvaraju2017grad} result shown in Figure \ref{fig:gradcam} as an example to illustrate our idea. Assuming that our machine has never seen the images of \textit{black-footed ferret} before. If ignoring label semantic information, the model can be easily misguided by background noise and large intra-class variations when learning with only one sample of \textit{black-footed ferret}, thus may take the striped background as the key feature of this category. On the contrary, our method (SEGA) can tell that \textit{black-footed ferret} is a kind of animal based on semantic knowledge, which guides our model to pay attention to key features when dealing with animals.

%-------------------------------------------------------------------------
\section{Related Works}

The mainstream few-shot learning methods utilize the auxiliary task to transfer visual prior knowledge from base classes to few-shot classes \cite{qi2020small, wang2020generalizing}. From the perspective of how to design auxiliary tasks, there exist common approaches such as metric-based, hallucination-based, and meta-learning-based. However, the above taxonomy cannot separate current methods well enough. Here we give a more orthogonal taxonomy from the perspective of how to use the support set. The first kind uses the support set to directly generate the classifiers for novel classes, which is related to lazy learning \cite{aha2013lazy}. It contains the metric-based methods \cite{snell2017prototypical,sung2018learning,vinyals2016matching, koch2015siamese,nickel2017poincar} and the classification weight generation methods \cite{qiao2018few,hou2019learning,gidaris2018dynamic,qi2018low}. The other kind uses the support set to finetune the end-to-end network, which is related to eager learning. It contains hallucination-based \cite{gao2018low,guan2020zero,schonfeld2019generalized,afrasiyabi2020associative} and optimization-based methods \cite{finn2017model,satorras2018few,ravi2016optimization,wang2019tafe,rusu2018meta}. Our approach belongs to the lazy learning category.

\textbf{Lazy learning methods.}
The first kind of lazy learning method is metric-based methods, whose core idea is to learn a metric space in which the samples of the same category are near each other while different categories are far away. This idea can be traced back to NCA \cite{goldberger2004neighbourhood} and LMNN \cite{weinberger2009distance}. Nowadays, there emerge many deep learning metric-based methods such as Matching Networks \cite{vinyals2016matching} which learns a nearest neighbor classifier with the help of context information, Prototypical Networks \cite{snell2017prototypical} which generates class prototype as the mean of support set samples, Relation Network \cite{sung2018learning} which uses the neural network to model the distance measurement, \etc. Another kind of lazy learning method is classification weight generation methods, which generate the class weight for novel classes directly \cite{qiao2018few,gidaris2018dynamic,qi2018low}. The cosine classifier is widely used to avoid the norm problem \cite{hou2019learning,qi2018low}. %However, the norm of classification weight is especially important for generalization which is already verified in transfer learning and incremental learning . Thus, the cosine classifier is proposed to deal with the norm problem \cite{gidaris2018dynamic,qi2018low}. With the normalization, the similarity calculation is only based on the angle, which can generalize well on novel classes.
Our approach lies in this classification weight generation category.

Most recently, there emerge works that use semantic knowledge in few-shot learning \cite{chen2019multi,schwartz2019baby,xing2019adaptive,peng2019few,li2019large} whose idea comes from a related research area named zero-shot learning \cite{farhadi2009describing,lampert2009learning,lampert2013attribute}. The source of semantic knowledge can be attributes, word embeddings, and even knowledge graphs.

\textbf{Semantic knowledge in few-shot learning.}
Thanks to the development in natural language processing, we can get label embeddings from the pre-training word embedding models such as GloVe \cite{pennington2014glove}. TriNet \cite{chen2019multi} takes the class label embedding to hallucinate new samples in semantic feature space. TRAML \cite{li2020boosting} uses class label embedding to generate adaptive margin loss. In addition, AM3 \cite{xing2019adaptive} uses label embedding to generate a semantic prototype which is used to perform a convex combination with the visual prototype to form the final class representation. Furthermore, MultiSem \cite{schwartz2019baby} introduces extra semantic like verbal descriptions and more recent work \cite{zhang2021prototype} extracts parts/attributes from WordNet \cite{miller1998wordnet}. Apart from semantic knowledge from language, correlation knowledge, which can be obtained from knowledge graph (\eg NEIL \cite{mitchell2018never}, WordNet \cite{miller1998wordnet}, \etc), can also be helpful in FSL. KTN \cite{peng2019few} proposes to construct a GCN in which the node representation comes from label embedding and the edge comes from knowledge graph to transfer knowledge from base class to novel class. KGTN \cite{chen2020knowledge} employs a similar idea to construct a GCN whose edge weights are generated by the semantic hierarchy of categories. 

As we can see, previous methods can be divided into two paradigms: semantic-dominated (more rely on semantic, \eg TriNet \cite{chen2019multi}, TRAML \cite{li2020boosting} \etc) and multimodal-fusion (fuse semantic and visual equally, \eg AM3 \cite{xing2019adaptive}, KTN  \cite{peng2019few} \etc). Here in this paper, we propose a new paradigm that is visual-dominated (\ie just use semantic to enhance, \eg our SEGA uses semantic just to generate visual attention instead of reconstruction). The advantages lie in (1) More robust and easier to learn since only need to learn attention while previous two paradigms overuse semantic to reconstruct visual information (\eg AM3 \cite{xing2019adaptive} reconstructs semantic prototype directly in visual space). (2) More reasonable since the essence of semantic is invariant (like attention in SEGA), we shouldn’t expect it useful in equivariant jobs (like generation/fusion in the previous two paradigms). (3) Customized for FSL since class-specific semantic attention can eliminate background noise and intra-class variations which are inevitably amplified in FSL.

%It becomes acceptable to use semantic knowledge for the lack of visual data in FSL. Previous methods can be divided into 2 paradigms: semantic-dominated (more rely on semantic, \eg TriNet, TRAML \etc) and multimodal-fusion (fuse semantic and visual equally, \eg AM3, KTN \etc).  \textbf{We propose a new paradigm that is visual-dominated (\ie just use semantic to enhance, \eg our SEGA)}. Advantages: (1) More robust and easier to learn since only need to learn attention while previous 2 paradigms overuse semantic to reconstruct visual information. (2) More reasonable since \textbf{the essence of semantic is INVARIANT (like attention in SEGA), we shouldn’t expect it useful in EQUIVARIANT jobs (like generation/fusion in previous 2 paradigms)}, as noted in \textcolor{magenta}{L142-150}. (3) Customized for FSL since class-specific semantic attention can eliminate background noise and intra-class variations (proved in \textcolor{magenta}{§4.3}) which are inevitably amplified in FSL while other paradigms fail to aim at this key challenge. As for framework, DynamicFSL-like framework (2 training stages and cosine classifier) is widely used in FSL and our key contribution is built upon it and beyond it.

%------------------------------------------------------------------------
\section{Approach}

We give some fundamental analysis for why semantic knowledge could be helpful in FSL in \S\ref{sec:why}; then we establish preliminaries about problem setting in \S\ref{sec:setting} and introduce general framwork in \S\ref{sec:frame}; finally, we focus on our SEGA in \S\ref{sec:sega} followed by some discussion in \S\ref{sec:discussion}.

\subsection{Why Utilize Semantic Knowledge?}
\label{sec:why} 

The introduction of semantic knowledge is nothing new in ZSL since the ZSL task cannot be completed without semantic knowledge. While in FSL previous works hardly use semantic knowledge. Most recently, there emerge some works that propose to utilize semantic knowledge in FSL \cite{chen2019multi,schwartz2019baby,xing2019adaptive,peng2019few}. However,
%to the best of our knowledge,%
there lacks of fundamental analysis about why using semantic knowledge can help in FSL. Thus, we adopt the Canonical Correlation Analysis (CCA) for aligning the visual and semantic features to the same latent space to analyze the correlation of visual space and semantic space. We train a CCA model on the visual features and word embeddings of 64 base classes $\mathcal{D}^{base}$ from miniImageNet, where visual features are obtained by the feature extractor trained on base classes. Then we perform the same CCA model on visual features and word embeddings of 16 validation classes $\mathcal{D}^{val}$ and 20 test classes $\mathcal{D}^{test}$ respectively to calculate the correlation coefficient. There is still a relatively high correlation between visual and semantic space on these novel classes while the correlation coefficient is quite small when using the non-corresponding visual and semantic data to train the CCA (results can be found in the supplementary material). We can draw a conclusion that visual space and semantic space are quite relevant and the alignment between them calculated on base classes can be transferred to novel classes.

\subsection{Problem Formulation}
\label{sec:setting}
Before testing on novel class, we are given \textit{M} base classes (denoted as $\mathcal{Y}^{b}$) for meta-learning purpose. During testing, there are \textit{N} novel classes (denoted as $\mathcal{Y}^{n}$) in each few-shot learning task where the base classes and novel classes are disjoint, $\ie$, $\mathcal{Y}^{b} \cap \mathcal{Y}^{n} = \emptyset$. We use the index $\{1, ...,M\}$ to represent the base classes and $\{M + 1, ...,M + N\}$ to represent the novel classes. The base classes dataset (denoted as $\mathcal{D}^{base}$) has plenty of samples per class, while the novel class dataset named support set (denoted as $\mathcal{D}^{novel}$) has only \textit{K} labeled samples per class. The support set contains $N \times K$ labeled images $\mathcal{D}^{novel} = \left\{\left(\boldsymbol{x}_{i}, y_{i}\right) \mid \boldsymbol{x}_{i} \in \mathcal{X}, y_{i} \in \mathcal{Y}^{n}\right\}_{i=1}^{N \times K}$, thus we call it the \textit{N}-Way, \textit{K}-Shot setting. $\mathcal{X} \in \mathbb{R}^{d_v}$ represents the visual space. To derive visual attention from class semantic knowledge, semantic information $\mathcal{S} = \{\boldsymbol{s}^c \in \mathbb{R}^{d_s}\}_{c=1}^{M+N}$ is provided for each class $c \in \mathcal{Y}^{b} \cup \mathcal{Y}^{n}$. The goal of FSL is to learn the classifiers for novel classes  $f_{fsl}: \mathcal{X} \rightarrow \mathcal{Y}^{n}$.

\begin{figure}
	\begin{center}
		%\fbox{\rule{0pt}{2in} \rule{.9\linewidth}{0pt}}
		\includegraphics[width=1.0\linewidth]{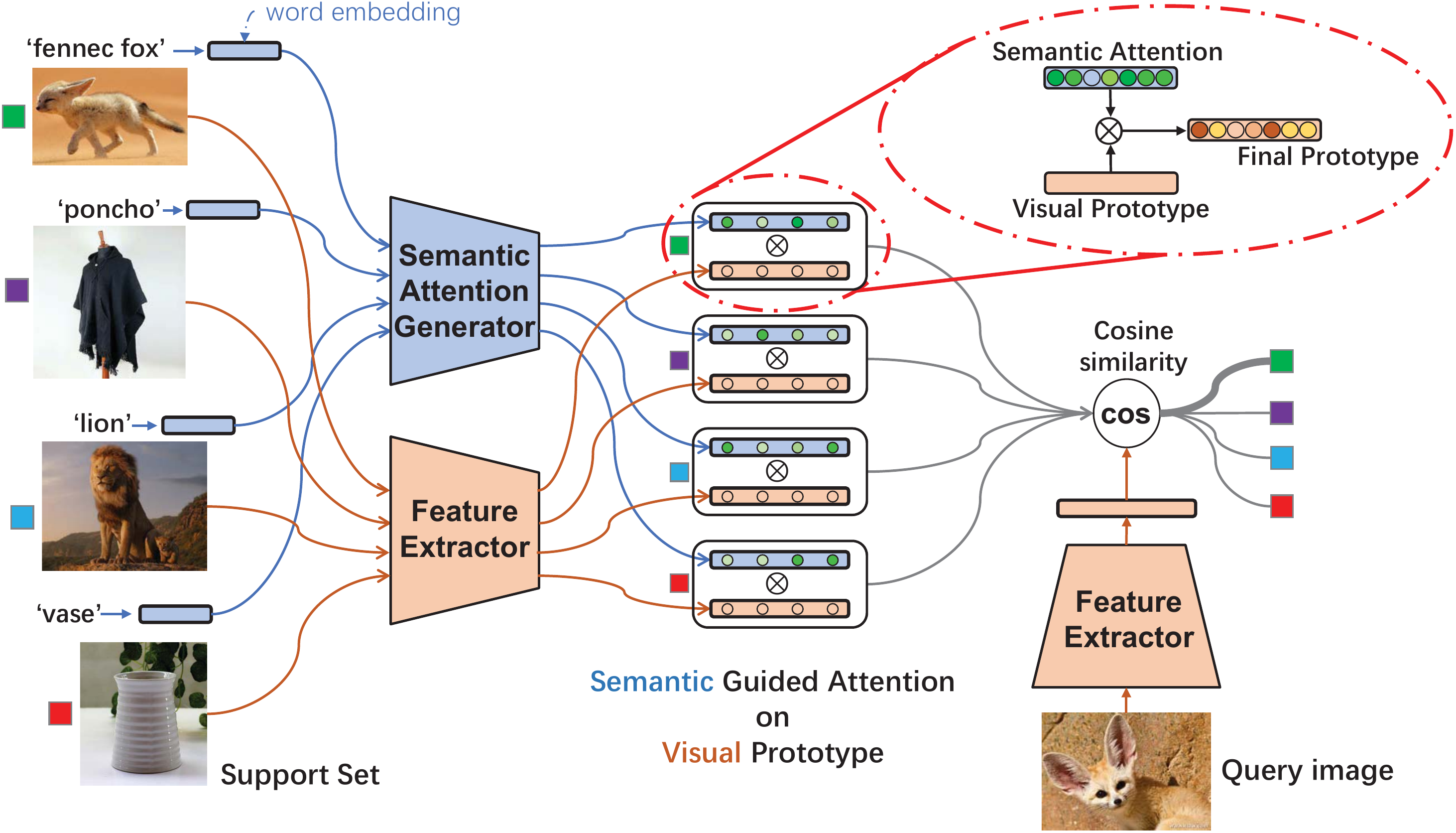}
	\end{center}
	\vspace{-0.45cm}
	\caption{The framework of our proposed Semantic Guided Attention method. We highlight the process of semantic guided attention which is to apply generated semantic attention to visual prototype by Hadamard product.  ``$\otimes$" denotes the Hadamard product operation and ``cos" denotes the cosine classifier.}
	\label{fig:framework}
	\vspace{-0.55cm}
\end{figure}

\subsection{Framework}
\label{sec:frame}
Figure \ref{fig:framework} shows the framework of our Semantic Guided Attention (SEGA). It contains three submodules as follows.

\textbf{Feature Extractor}. 
Just like that a human cannot learn a novel category without having seen anything before, the FSL method cannot learn novel class $\mathcal{D}^{novel}$ without the help of base classes $\mathcal{D}^{base}$. The base classes $\mathcal{D}^{base}$ are used to train the \textit{Feature Extractor} in Figure \ref{fig:framework}. The training procedure follows the classical deep learning paradigm with the common backbone. After training we fix the \textit{Feature Extractor} just like many other works, and now we get the visual space $\mathcal{X} \in \mathbb{R}^{d_v}$. Without doubt, the performance of visual space $\mathcal{X}$ significantly depends on the backbone capacity, the number of classes in $\mathcal{D}^{base}$, and the quality and diversity of samples in $\mathcal{D}^{base}$. So it is worth knowing and keeping all settings the same for experimental study later. 

\textbf{Cosine Classifier}.
Apart from feature extractor, the general deep learning framework consists of a classifier. The standard classifier employs dot-product to calculate classification scores. It is not suitable for FSL setting since the norm of generated weight is not controllable. Therefore during few-shot training we use \textit{Cosine Classifier} proposed by \cite{gidaris2018dynamic,qi2018low} in which the classification score is calculated based on cosine similarity $\text {score}_{k}(\boldsymbol{x})=\text {cos}\left\langle {\boldsymbol{x}, \boldsymbol{w}}_{k}\right\rangle=\left \langle \frac{\boldsymbol{x}}{\| \boldsymbol{x} \|}, \frac{{\boldsymbol{w}}_{k}}{\| {\boldsymbol{w}}_{k} \|} \right \rangle$. With the normalization, the similarity calculation is only based on angle and can generalize well. Besides, the classifier weight is equivalent to class prototype in metric learning when using cosine classifier since the optimization target is mathematically equivalent \cite{qi2018low}.

\textbf{Classification Weight Generator}.
The classification weights can be tuned adequately when given plenty of images per class. However, there are not enough samples to tune the classification weight in FSL. As noted before, our method follows the weight generation paradigm which means we will generate the classification weights $\mathcal{W}^{novel} = \{\boldsymbol{w}_c\}_{c=M+1}^{M+N}$ for each novel class. Specifically, our classification weight generator is a semantic guided attention weight generator which will be elaborated later.

\textbf{Training Procedure}.
To fully mimic human learning process, the training procedure consists of two stages. The first stage is to train the \textit{Feature Extractor} on the base classes $\mathcal{D}^{base}$, which follows the standard classification training paradigm without any semantic knowledge or few-shot concerns. After the first training stage, the \textit{Feature Extractor} will be fixed. The second stage, \ie few-shot training, is of vital importance which contains the training of the \textit{Semantic Guided Attention Weight Generator} and \textit{Cosine Classifier}. Here we adopt the strategy in Dynamic-FSL \cite{gidaris2018dynamic} whose idea is similar with \textit{episodic training} \cite{vinyals2016matching}. Compared to \textit{episodic training}, the difference is that our strategy performs the classification task across the whole base classes $\mathcal{Y}^{b}$ and simulates testing scenario at the same time. More specifically, for each episode, we randomly sample $N$ classes from the base classes $\mathcal{Y}^{b}$ to act as ``novel" classes, then sample $K$ samples from each ``novel" class to form a fake \textit{N}-Way \textit{K}-Shot support set. As shown in Figure \ref{fig:framework}, we can calculate $N$ visual prototypes and enhance them using semantic guided attentions. Thus we get $N$ classification weights which are used to replace the corresponding base classification weights (other weights are also enhanced by their own semantic attentions) in \textit{Cosine Classifier}, and then perform classification and cross-entropy loss calculation.

\subsection{Semantic Guided Attention Weight Generator}
\label{sec:sega}
Two key components in our classification weight generation are \textit{Visual Prototype} and \textit{Semantic Guided Attention}. As shown in Figure \ref{fig:framework}, the semantic guided attention will be applied to enhance the visual prototype and the result will be set as the final classification weight.

\textbf{Visual Prototype.}
As noted above, the classifier weight is equivalent to the class prototype in metric learning when using the cosine classifier. Hence, we generate the classification weight based on support sets $\mathcal{D}^{novel}$ just like most other metric-based FSL methods. Following the classical Prototypical Network \cite{snell2017prototypical}, we get the visual prototype:
\begin{equation}
	\mathbf{p}_{avg}^{c}=\frac{1}{\left|\mathcal{D}^{n}_c\right|} \sum_{\left(\boldsymbol{x_{i}}, y_{i}\right) \in \mathcal{D}^{n}_c} \boldsymbol{x_{i}},
\end{equation}
where $\mathcal{D}^{n}_c \in \mathcal{D}^{novel}$ is the subset of the support set which contains the samples belonging to class \textit{c}.
Without doubt, this prototype generation way is too straightforward which ignores the visual prior knowledge that can be transferred from the base class weights $\mathcal{W}^{base} = \{\boldsymbol{w}_c\}_{c=1}^{M}$. Therefore, we follow our baseline method \cite{gidaris2018dynamic} to transfer the visual prior from base class weights based on the cosine similarity to enhance the averaging prototype:
\begin{equation}
	\mathbf{p}_{att}^{c}=\frac{1}{\left|\mathcal{D}^{n}_c\right|} \sum_{\left(\boldsymbol{x_{i}}, y_{i}\right) \in \mathcal{D}^{n}_c} \sum_{j \in \mathcal{Y}^{b}} Att\left(\boldsymbol{\phi_{q}} \boldsymbol{x_{i}}, \boldsymbol{k_{j}}\right) \cdot \boldsymbol{w}_j,
	\label{con:visual_prior}
\end{equation}
where $\boldsymbol{\phi_{q}} \in \mathbb{R}^{d_v \times d_v}$ is a learnable weight matrix and $\left\{\boldsymbol{k_{j}} \in \mathbb{R}^{d_v}\right\}_{j=1}^{M}$ is a set of $M$ learnable keys. $\boldsymbol{\phi_{q}}$ transforms the feature $\boldsymbol{x_{i}}$ to query vector which will be used to perform attention with $\boldsymbol{k_{j}}$ by a cosine based attention kernel $Att\left(\cdot,\cdot\right)$.
%where $\boldsymbol{\phi_{q}} \in \mathbb{R}^{d_v \times d_v}$ is a learnable weight matrix that transforms the feature $\boldsymbol{x_{i}}$ to query vector, which will be used to perform attention with $\left\{\boldsymbol{k_{j}} \in \mathbb{R}^{d_v}\right\}_{j=1}^{M}$ which is a set of $M$ learnable keys by a cosine based attention kernel $Att\left(\cdot,\cdot\right)$ which specifically is a cosine similarity function followed by a softmax over $M$ base classes.
By transferring visual prior knowledge from base classes, we model the final visual prototype as the combination of $\mathbf{p}_{avg}^{c}$ and $\mathbf{p}_{att}^{c}$:
\begin{equation}
	\mathbf{p}^{c} = \lambda_1 \times \mathbf{p}_{avg}^{c} + \lambda_2 \times \mathbf{p}_{att}^{c},
\end{equation}
where $\lambda_1, \lambda_2 \in \mathbb{R}$ are learnable coefficients. 
The final visual prototype will then be feature-selected by semantic guided attention as introduced next.

\textbf{Semantic Guided Attention.}
The visual prototype is neither precise nor stable on account of the lack of image samples, thus we propose to use the semantic knowledge to guide the attention on the visual prototype in top-down manner as shown in Figure \ref{fig:framework}. The semantic knowledge can come from the class labels, attributes, and even knowledge graph. Here we choose the word embeddings of class labels as the semantic knowledge source: $\mathcal{S} = \{\boldsymbol{s}^c \in \mathbb{R}^{d_s}\}_{c=1}^{M+N}$, where $\boldsymbol{s}^c$ is the word embedding of the label of class \textit{c}. $d_s$ is the dimension of the word embedding space. We use an MLP to model the transformation $g: \mathbb{R}^{d_s} \rightarrow \mathbb{R}^{d_v}$ which maps the word embedding in semantic space $\mathbb{R}^{d_s}$ to the visual attention  in visual space $\mathcal{X}$: $\boldsymbol{a}_c = g(\boldsymbol{s}^c)$. The last layer of $g$ is a sigmoid function, therefore the visual attention $\boldsymbol{a}_c$ is bounded between [0,1]. Actually, $\boldsymbol{a}_c$ can be understood as a feature selection in visual space $\mathcal{X}$ which selects the vital feature dimensions with respect to class \textit{c} guided by semantic knowledge $\boldsymbol{s}^c$, and the final classification weight for class \textit{c} is
\begin{equation}
	\boldsymbol{w}_{c} = \boldsymbol{a}_c \otimes \mathbf{p}^{c},
	\label{con:sem_att}
\end{equation}
where $\otimes$ denotes the Hadamard product (\ie, element-wise product operation). It is worth noting that we use \textit{Cosine Classifier} to perform final classification so that the classification score is calculated by 
\begin{equation}
	\text {score}_{c}(\boldsymbol{x})=t \cdot \text {cos}\left\langle {\boldsymbol{x}, \boldsymbol{w}_{c}}\right\rangle=t \cdot \text {cos}\left\langle {\boldsymbol{x}, \boldsymbol{a}_c \otimes \mathbf{p}^{c}}\right\rangle,
\end{equation}
where $t$ is the temperature coefficient to scale the cosine similarity in order to be better suitable for softmax. As we can see, the visual attention $\boldsymbol{a}_c$ here is also playing a role in transferring knowledge since the more similar the visual attention is, the more similar the classification weight will be. Following the same paradigm we can generate all $N$ novel classification weights $\{\boldsymbol{w}_{c}\}_{c=M+1}^{M+N}$, then we get the desired few-shot learning classifier $f_{fsl}$.

\subsection{Discussion}
\label{sec:discussion}
\textbf{Difference from the baseline.}
Our SEGA is mainly inspired by Dynamic-FSL \cite{gidaris2018dynamic} which is a widely-used framework in FSL proposing the cosine classifier and the classification weight generator. However, \cite{gidaris2018dynamic} doesn't involve semantic information which can play a critical role especially when there is a lack of visual experience. In contrast, our SEGA not only utilizes semantic but also works in a new paradigm that is visual-dominated, while previous methods are either semantic-dominated or multimodal-fusion. % to guide the classification weight generation. It is also worth noticing the difference between SEGA and AM3 \cite{xing2019adaptive}. SEGA and AM3 both map the class label embedding into the visual space $\mathcal{X}$. AM3 gets the semantic prototype while we get the semantic guided attention $\boldsymbol{a}_c$ which is bounded between [0,1]. Referring to the cognitive neurosciences and contrastive learning, we argue that the attention way is more human-like and robust since humans get the key discriminative features instead of a complete visual reconstruction from semantic knowledge.

\textbf{Further improvements.}
As we can see, our approach SEGA is orthogonal to most of the current FSL methods since we introduce semantic knowledge which is another dimension to enhance the visual prototype. That means SEGA can be combined with most unimodal metric-based FSL methods to get the prototype more stable and precise, especially in the case of extremely short of images like in 1-Shot learning scenario. On the other hand, semantic knowledge can come from many other sources. With the development of NLP, the semantic guidance imposed on visual features would be more accurate by exploring more powerful knowledge sources such as visual knowledge base, BERT embedding and so on.

%------------------------------------------------------------------------
\section{Experiments}
In this section, we evaluate our method on four benchmark datasets and then analyze the effectiveness of it.

\subsection{Datasets and Settings}

\begin{table*}[]
	\caption{Ablation study of our proposed method on miniImageNet, tieredImageNet, and CIFAR-FS. We report the average classification accuracies (\%) on 5000 test episodes of novel categories (with 95\% confidence intervals). ``Sem." denotes whether to use semantic defined in Equation(\ref{con:sem_att}) and ``FAKE" means using the non-corresponding label as semantic guidance.}
	\label{table:ablation_study}
	\vspace{-0.55cm}
	\begin{center}
		\begin{tabular}{c|llllll}
			\hline
			\multirow{2}{*}{Sem.}     & \multicolumn{2}{c}{miniImageNet}                                                              & \multicolumn{2}{c}{tieredImageNet}                                                            & \multicolumn{2}{c}{CIFAR-FS}                                                                  \\
			& \multicolumn{1}{c}{5Way 1Shot}                & \multicolumn{1}{c}{10Way 1Shot}               & \multicolumn{1}{c}{5Way 1Shot}                & \multicolumn{1}{c}{10Way 1Shot}               & \multicolumn{1}{c}{5Way 1Shot}                & \multicolumn{1}{c}{10Way 1Shot}               \\ \hline
			YES                       & \textbf{69.04}$\pm$0.26\textbf{($\uparrow$6)} & \textbf{52.71}$\pm$0.15\textbf{($\uparrow$6)} & \textbf{72.18}$\pm$0.30\textbf{($\uparrow$4)} & \textbf{56.82}$\pm$0.21\textbf{($\uparrow$3)} & \textbf{76.24}$\pm$0.25\textbf{($\uparrow$8)} & \textbf{61.77}$\pm$0.17\textbf{($\uparrow$8)} \\
			NO                        & 62.81$\pm$0.27{( -- )}                        & 46.73$\pm$0.17{( -- )}                        & 68.55$\pm$0.31{( -- )}                        & $54.01\pm$0.21{( -- )}                        & 67.78$\pm$0.30{( -- )}                        & 53.32$\pm$0.21{( -- )}                        \\
			\multicolumn{1}{l|}{FAKE} & 59.04$\pm$0.27{($\downarrow$4)}               & 43.58$\pm$0.16{($\downarrow$3)}               & 64.64$\pm$0.31{($\downarrow$4)}               & 50.07$\pm$0.21{($\downarrow$4)}               & 63.27$\pm$0.29{($\downarrow$4)}               & 48.66$\pm$0.19{($\downarrow$5)}               \\ \hline
		\end{tabular}
	\end{center}
	\vspace{-0.7cm}
\end{table*}

\textbf{Datasets.}
We perform experiments on four widely used FSL benchmarks to verify the effectiveness of the proposed method, $\ie$, miniImageNet \cite{vinyals2016matching}, tieredImageNet \cite{ren2018meta}, CIFAR-FS \cite{bertinetto2019meta}, and CUB \cite{wah2011caltech}. miniImageNet and tieredImageNet are both derivatives of ImageNet dataset \cite{russakovsky2015imagenet}, CIFAR-FS is derived from CIFAR-100 dataset \cite{krizhevsky2009learning, torralba200880}. The datasets summary can be found in supplementary material.

\textbf{Semantic knowledge source.}
We use GloVe \cite{pennington2014glove} to be our semantic knowledge source which is a word embedding model trained on the Wikipedia dataset and the dimension of the word embedding is 300. We use it to get word embedding for each category. When the category label contains more than one word (\eg ``baseball bat"), most previous methods will generate the embedding by averaging them (\eg $\frac{emb(``baseball")+emb(``bat")}{2}$) which is unreasonable. Besides, most previous methods use a 300-dimension zero vector instead when there is no annotation found in GloVe's vocabulary. Our method uses the hypernym synset based on WordNet to deal with these problems. For those annotations not found in GloVe, we refer to its hypernym synset in WordNet until there is an annotation found in GloVe.  By this way, more accurate semantic information can be explored for each category. 

\textbf{Implementation details.}
All experiments are conducted under in PyTorch framework\footnote{{The codes are at \url{http://vipl.ict.ac.cn/resources/codes} or \url{https://github.com/MartaYang/SEGA}}}. For all datasets, we utilize a ResNet-12 as our backbone following most previous works\cite{chen2019multi, gidaris2019generating, li2020boosting, liu2020negative, xing2019adaptive}. We also change the number of filters from (64,128,256,512) to (64,160,320,640) same as \cite{lee2019meta, ravichandran2019few, tian2020rethinking}. To avoid overfitting we follow most prior works \cite{liu2020negative, zhang2020deepemd,hou2019cross, wang2020cooperative} to adopt the random crop, color jittering, erasing and Dropblock \cite{ghiasi2018dropblock} regularization. The semantic guided attention generator used in all cases is an MLP, with 2 fully connected layers and a dropout layer between them, followed by sigmoid nonlinearity. Other parameters $\lambda_1$, $\lambda_2$, and cosine similarity temperature $t$ are tuned during the training of the generator. We use SGD optimizer with a momentum of 0.9 and weight decay of 5e-4. During the first training stage, we train the \textit{Feature Extractor} for 60 epochs (90 for tieredImageNet), with each epoch consisting of 1000 episodes. As for the second training stage, we train \textit{Semantic Guided Attention Weight Generator} and \textit{Cosine Classifier} for 20 epochs in all cases. We adopt an empirical learning rate scheduler following the practice of \cite{gidaris2018dynamic, lee2019meta, wang2020cooperative}. More details can be found in the supplementary material.

\begin{figure*}[t]
	\centering
	\subfigure[Before semantic attention]{
		\label{fig:proto_change:a} %% label for first subfigure
		\includegraphics[width=5.4cm,height=3.6cm]{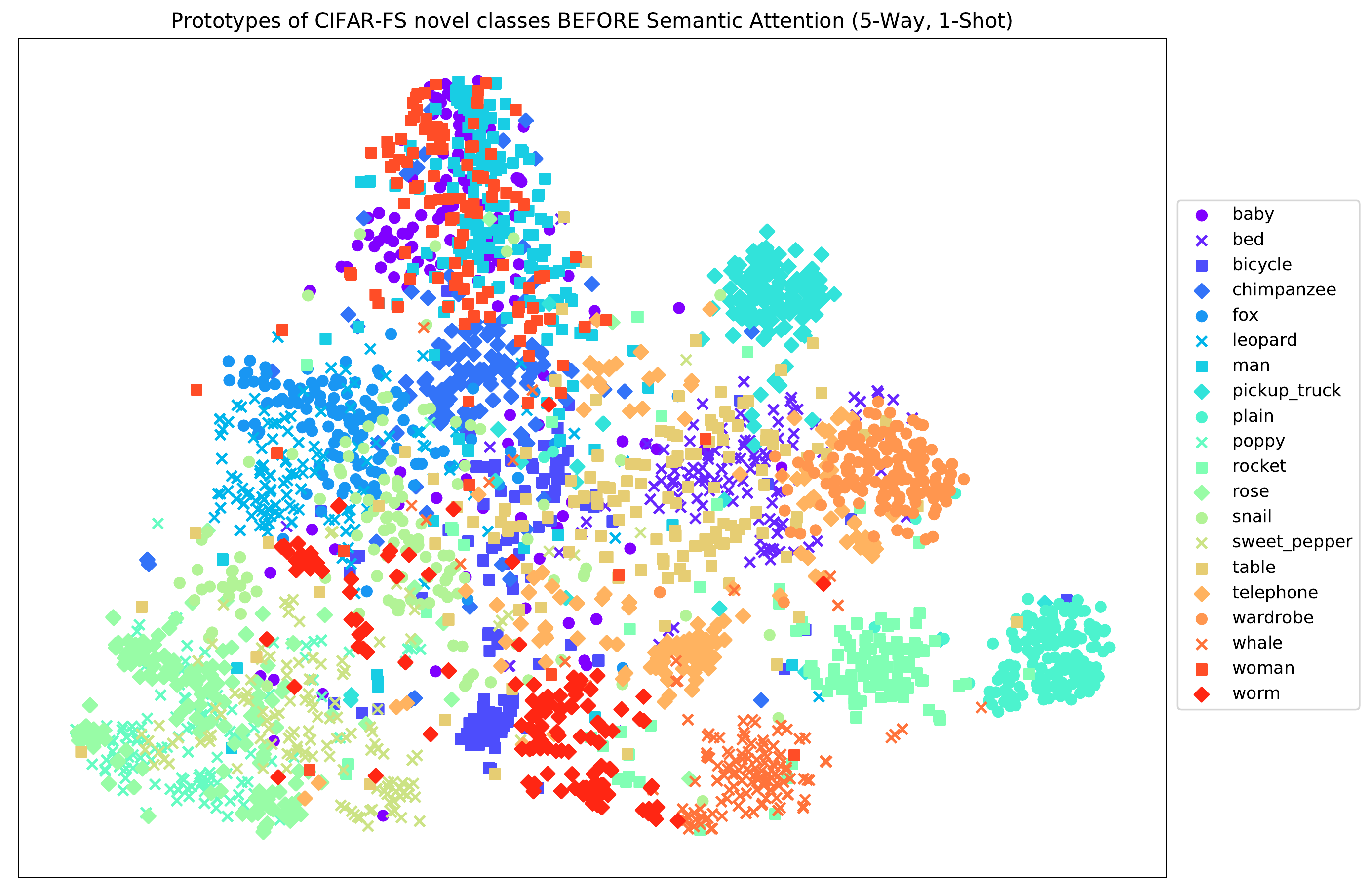}}
	\hspace{0.05in}
	\subfigure[After semantic attention]{
		\label{fig:proto_change:b} %% label for second subfigure
		\includegraphics[width=5.4cm,height=3.6cm]{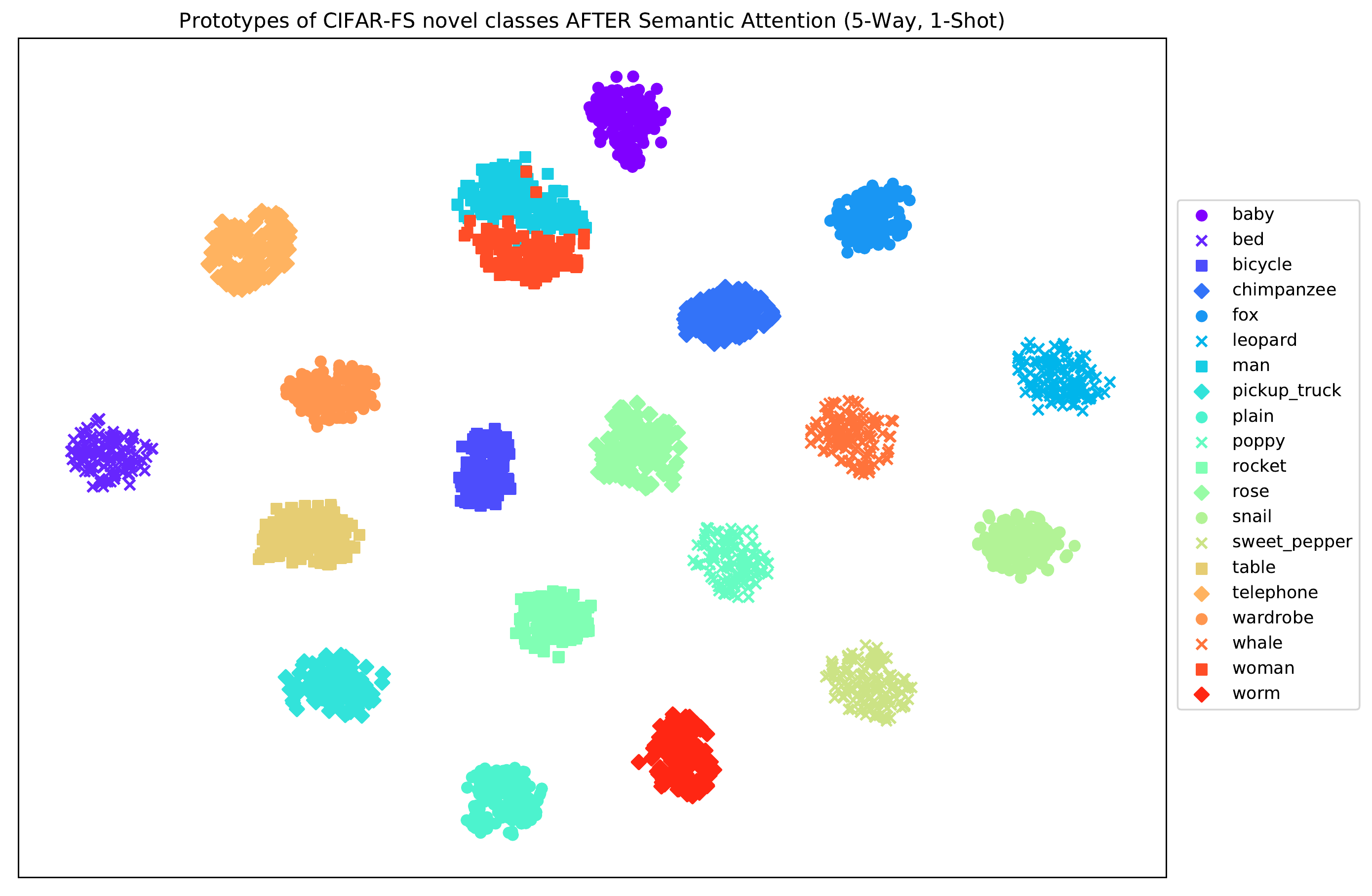}}
	\hspace{0.05in}
	\subfigure[After \textbf{inversed} semantic attention]{
		\label{fig:proto_change:c} %% label for second subfigure
		\includegraphics[width=5.4cm,height=3.6cm]{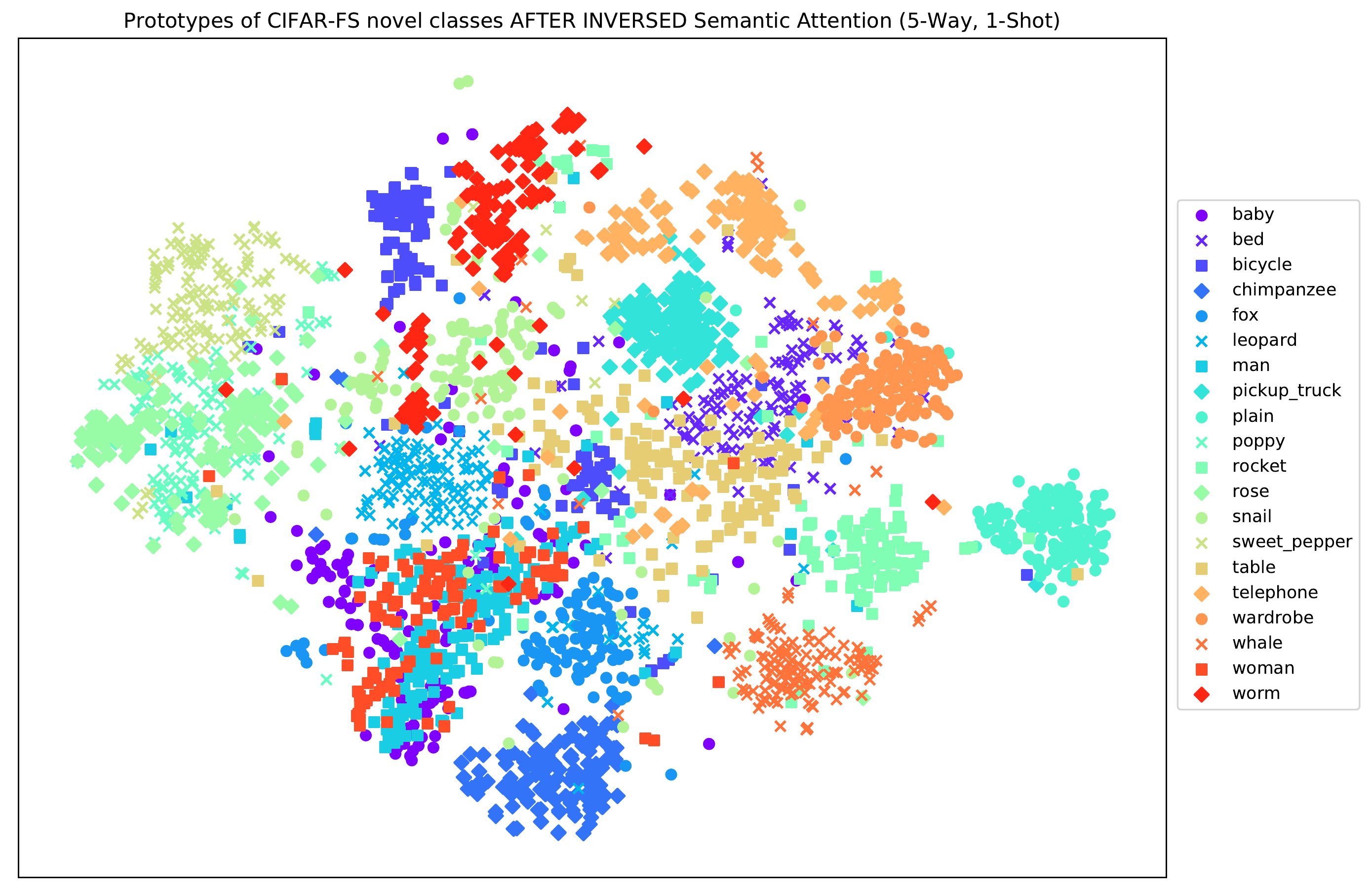}}
	\vspace{-0.25cm}
	\caption{t-SNE visualization of the prototypes in visual space under 5-Way 1-Shot scenario. \textbf{(a)} and \textbf{(b)} are prototypes before ($\mathbf{p}^{c}$) and after ($\boldsymbol{a}_c \otimes \mathbf{p}^{c}$) performing the semantic attention. \textbf{(c)} shows the result when applying the \textbf{}inverse attention ($(1-\boldsymbol{a}_c) \otimes \mathbf{p}^{c}$). Note that the prototypes are all generated during 600 epochs testing on 20 unseen novel classes on CIFAR-FS (results for miniImageNet can be found in supplementary material) and the point color represents its category. All the prototypes are L2-normalized since we use the cosine classifier.}
	\label{fig:proto_change} %% label for entire figure
	\vspace{-0.45cm}
\end{figure*}

\subsection{Effectiveness of the Proposed Framework}
%In this subsection, we verify the effectiveness of our method and give some qualitative results to further strengthen the conclusion. 

As shown in Table \ref{table:ablation_study}, we conduct ablation studies to verify the effectiveness of the proposed semantic guided attention weight generator. By comparing the performance of with semantic (the first row) and without semantic (the second row which is our baseline \cite{gidaris2018dynamic} under our framework), we can infer that the semantic knowledge can significantly improve performance (\ie, the performance improvement is 6\%, 4\% and 8\% on miniImageNet, tieredImageNet, and CIFAR-FS respectively). It is also worth noting that the correct guidance is rather important since when using fake semantic knowledge (the third row), which means using the irrelevant semantic label to generate class attention, the performance drops even lower than the result without semantic guidance. Furthermore, we conduct experiments on different kinds of pre-trained word embedding models like Word2Vec \cite{mikolov2013efficient} and see similar phenomena which can be found in the supplementary material.
%\textbf{Visual Prior}: From Table \ref{table:ablation_study}, we can conclude that it is also very useful to transfer visual prior knowledge from base class weights especially when using semantic guidance. The explanation is that the semantic attention network is trained on the base class, thus when dealing with a novel category the semantic guided attention will be more accurate after transferring the visual prior of most similar base category which the network knows where to pay attention to.

\subsection{Dive Deep into Semantic Attention}
\textbf{What does semantic attention do?} To better understand how the semantic attention works, we perform the t-SNE visualization. Figure \ref{fig:proto_change:a} and \ref{fig:proto_change:b} shows the change of the prototypes before and after applying the semantic attention under 5-Way 1-Shot scenario. As we can see, before the semantic attention the generated prototypes $ \mathbf{p}^{c} $ are quite unstable, but after applying semantic attention the final prototypes $ \boldsymbol{a}_c \otimes \mathbf{p}^{c} $ become a lot more stable (\ie, the prototypes of the same class get closer and vice versa). It gives the reason why our model can get significant gain under 1-Shot setting. We also show the results when applying inversed semantic guided attention $(1-\boldsymbol{a}_c) \otimes \mathbf{p}^{c}$ in Figure \ref{fig:proto_change:c} which means to ignore dimensions that our model thinks is important while emphasizing the unimportant ones. After that, the prototypes get even more unstable and chaotic which further demonstrates that our SEGA does capture the class-specific discriminative dimensions.

\textbf{Where does SEGA pay attention to?} Figure \ref{fig:gradcam} shows the Grad-CAM \cite{selvaraju2017grad} visualization testing on miniImageNet unseen classes based on our model with ResNet-12 backbone. Noted that the model remains exactly the same for all three results columns, the only difference is the attention vector to be applied. By comparing the results of “SEGA” ($ \boldsymbol{w}_{c} = \boldsymbol{a}_c \otimes \mathbf{p}^{c} $), “No\_Att” ($\boldsymbol{w}_{c} = \mathbf{p}^{c} $) and “Inv\_Att” ($\boldsymbol{w}_{c} = (1-\boldsymbol{a}_c) \otimes \mathbf{p}^{c}$), our SEGA can pay attention to the most crucial class-specific feature instead of the misguiding background noise and large intra-class variations. When only given one sample, if we do not utilize the semantic knowledge from the class label(\eg, \emph{bookshop}), even we human can get confused about what this category exactly is (\eg, when seeing a person in the bookshop). Thus, leveraging semantic knowledge from class labels is of vital importance, from which our model knows it should pay attention to the most crucial feature (\eg, books and bookshelf) instead of the noise (\eg, person) in the image.

\begin{figure}[t]
	\centering
	\subfigure[Attribute from Att. Intersection]{
		\label{fig:cam:a} %% label for second subfigure
		\includegraphics[width=3.9cm,height=3.9cm]{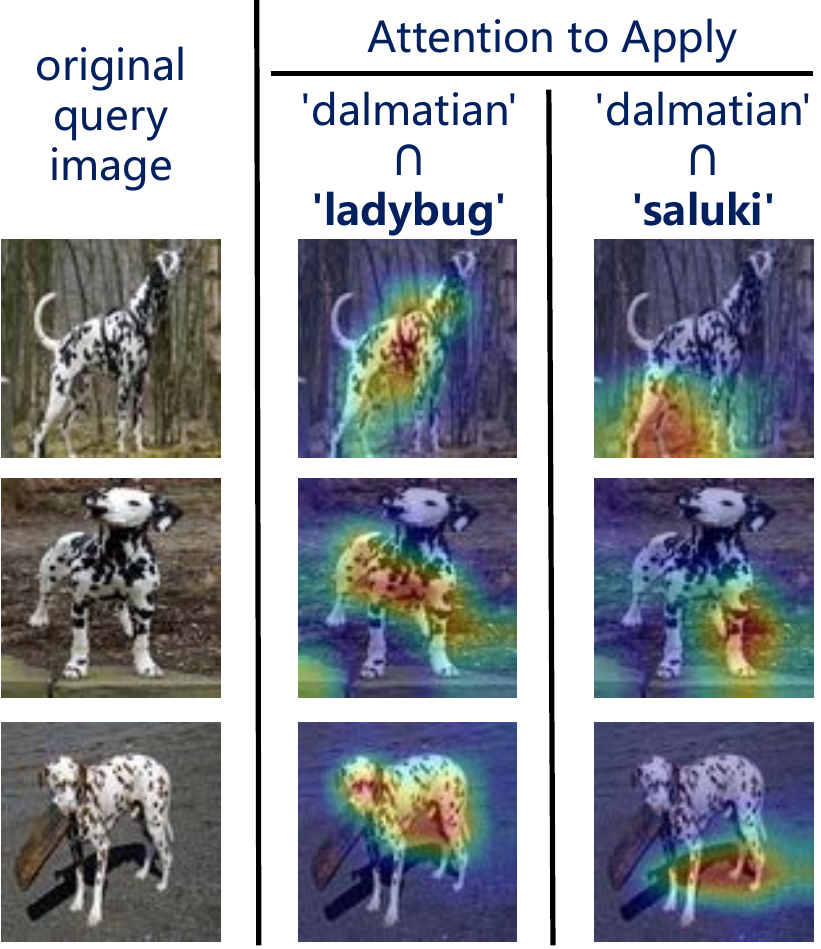}}
	\hspace{0.05in}
	\subfigure[CutMix Image with Each Att.]{
		\label{fig:cam:b} %% label for second subfigure
		\includegraphics[width=3.9cm,height=3.9cm]{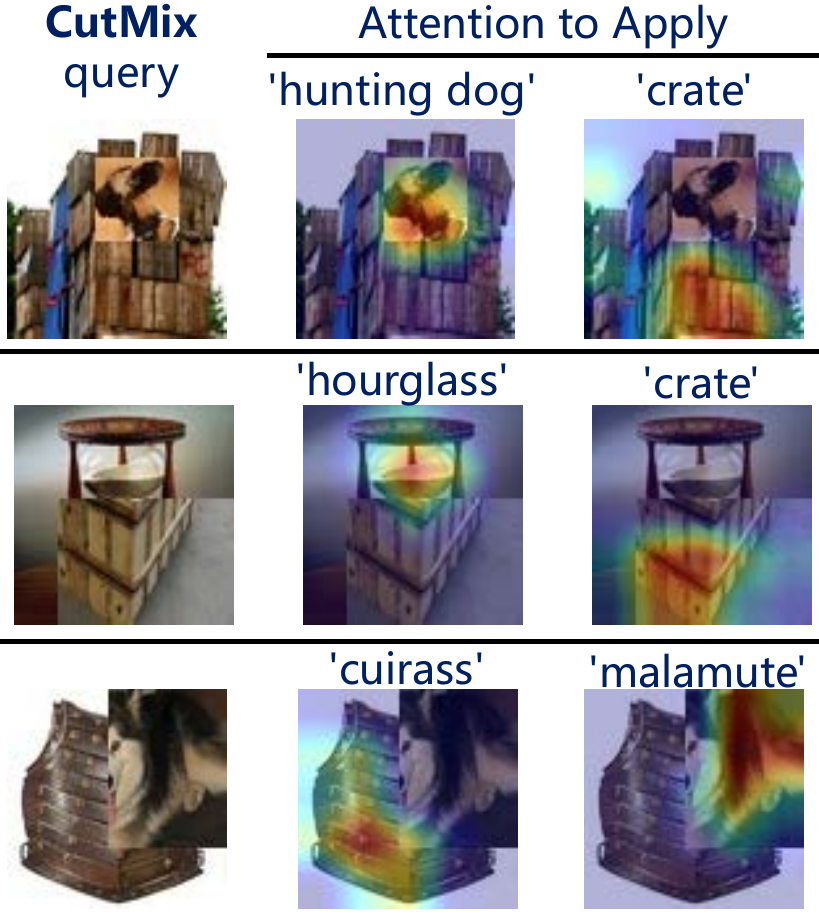}} \\
	\vspace{-0.35cm}
	\caption{\textbf{(a)} shows that by applying the intersection of two categories' attentions, the common ground attribute of these two category can be highlighted (\eg ``spots" for \textit{dalmatian} $\cap$ \textit{ladybug} and ``long legs" for \textit{dalmatian} $\cap$ \textit{saluki} ). \textbf{(b)} gives the results on the same CutMix query image when applying the attention of each two categories respectively.}
	\label{fig:cam} %% label for entire figure
	\vspace{-0.35cm}
\end{figure}

Furthermore, Figure \ref{fig:cam:b} is a harder task where the query is the CutMix \cite{yun2019cutmix} of two novel categories. Even in this circumstance, our SEGA can still pay attention to the key part of each corresponding category, which demonstrates the robustness of the attention generated by our model.

More interestingly, since our SEGA can generate class-specific attention, why not apply the intersection of two categories’ attentions to get their common ground attributes? Figure \ref{fig:cam:a} shows that common attribute “spots” will be highlighted when applying the intersection of attentions of \textit{dalmatian} and \textit{ladybug} (both of them have spots on body), and “long legs” will be highlight when applying \textit{dalmatian} $\cap$ \textit{saluki} (both of them have long legs). It suggests that our SEGA implicitly establishes a correspondence between semantic knowledge and visual attributes.

\begin{figure}
    \vspace{-0.05cm}
	\centering
	\subfigure[Hierarchical-Clustered Heatmap]{
		\label{fig:att_matrix:a} %% label for second subfigure
		\begin{minipage}[]{.5\linewidth}
			\centering
			\includegraphics[width=3.6cm,height=3.2cm]{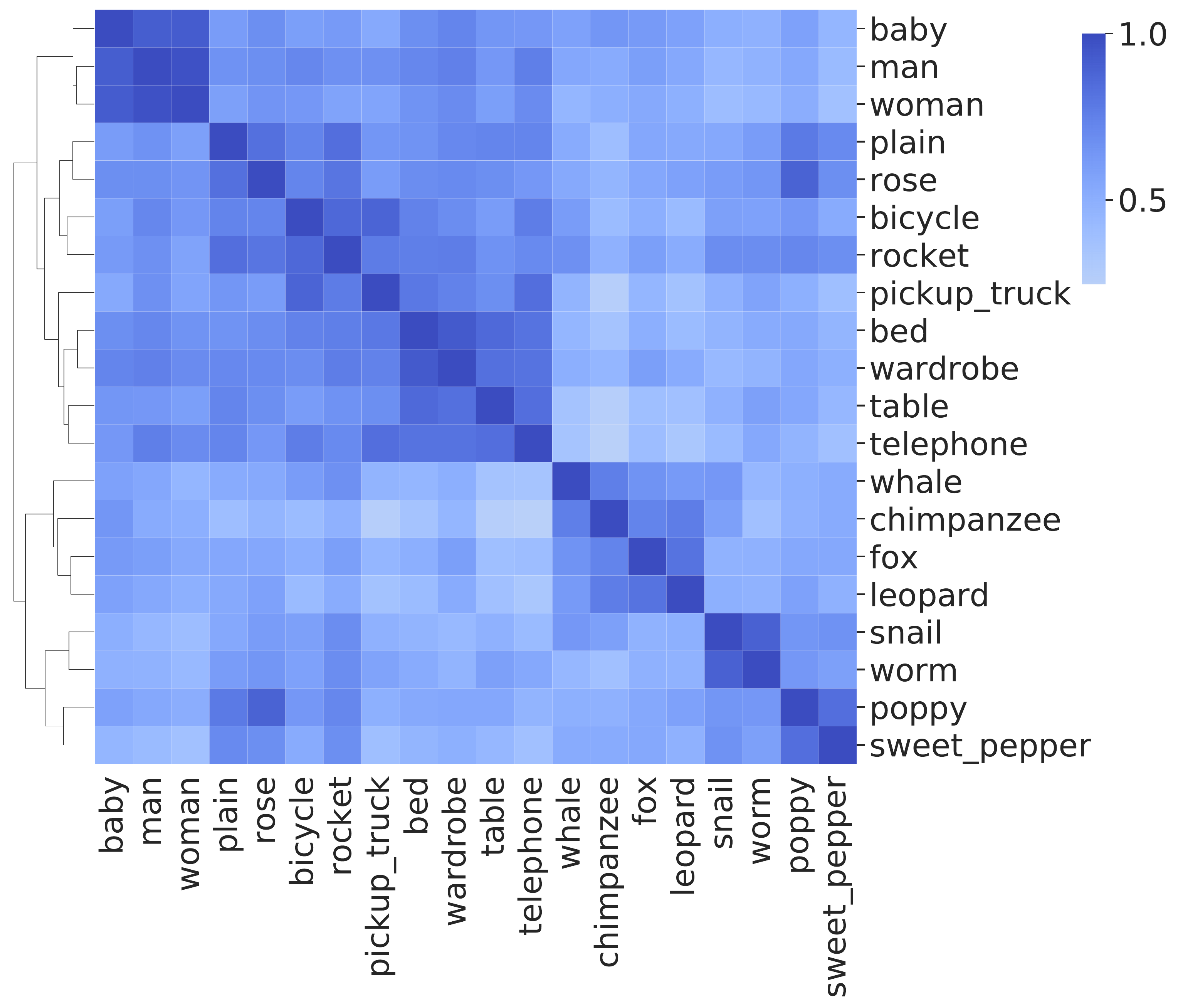}
			%\caption{Similarity matrix of attentions and the hierarchical clustering result}
	\end{minipage}}\noindent
	\subfigure[WordNet Hierarchical Structure]{
		\label{fig:att_matrix:b} %% label for second subfigure
		\begin{minipage}[]{.5\linewidth}
			\centering
			\resizebox{0.8\linewidth}{!}{
				\begin{tabular}{|c|c|c|c|}
					\hline
					\multirow{19}{*}{object} & \multirow{3}{*}{person}          & offspring    & \textbf{baby}                                                                       \\ \cline{3-4} 
					&                                  & adult        & \begin{tabular}[c]{@{}c@{}}\textbf{man}\\ \textbf{woman}\end{tabular}                        \\ \cline{2-4} 
					& \multirow{5}{*}{instrumentality} & vehicle      & \begin{tabular}[c]{@{}c@{}}\textbf{bicycle}\\ \textbf{rocket}\\ \textbf{pickup truck}\end{tabular}   \\ \cline{3-4} 
					&                                  & furniture    & \begin{tabular}[c]{@{}c@{}}\textbf{bed}\\ \textbf{wardrobe}\\ \textbf{table}\end{tabular}             \\ \cline{3-4} 
					&                                  & equipment    & \textbf{telephone}                                                                  \\ \cline{2-4} 
					& \multirow{4}{*}{animal}          & placental    & \begin{tabular}[c]{@{}c@{}}\textbf{whale}\\ \textbf{chimpanzee}\\ \textbf{fox}\\ \textbf{leopard}\end{tabular} \\ \cline{3-4} 
					&                                  & invertebrate & \begin{tabular}[c]{@{}c@{}}\textbf{snail}\\ \textbf{worm}\end{tabular}                       \\ \cline{2-4} 
					& \multirow{3}{*}{plant}           & angiosperm   & \textbf{poppy}                                                                      \\ \cline{3-4} 
					&                                  & shrub        & \begin{tabular}[c]{@{}c@{}}\textbf{rose}\\ \textbf{sweet pepper}\end{tabular}               \\ \hline
			\end{tabular}}
			%\captionof{table}{The hierarchical structure from WordNet}	
	\end{minipage}}
	\vspace{-0.45cm}
	\caption{In \textbf{(a)}, the attention vector for each class is generated by our attention generation model given novel class names of CIFAR-FS test set. Pearson correlation coefficient and the average-linkage algorithm are used for similarity matrix calculating and hierarchical clustering. \textbf{(b)} shows their ground truth hierarchical structure on WordNet, which can be basically matched with the former clustering result.}
	\label{fig:att_matrix} %% label for entire figure
	\vspace{-0.65cm}
\end{figure}

\textbf{Why does semantic attention work?} Figure \ref{fig:att_matrix:a} shows the similarity matrix of the generated attention vectors and their hierarchical clustering tree. It makes sense that visually similar categories cluster together and we can find block-diagonal phenomenon in the similarity matrix (\eg, the attention of baby, man, and woman cluster together and bicycle, rocket, and truck also cluster together). Note that even we do not explicitly exploit WordNet \cite{miller1998wordnet} knowledge database, our hierarchical clustering result is quite similar to the ground truth hierarchical structure in WordNet shown in Figure \ref{fig:att_matrix:b}, which again verifies the effectiveness of SEGA.

\begin{table*}[ht]
	\caption{Comparisons with popular FSL approaches in average classification accuracies (\%) on miniImageNet and tieredImageNet. We report the average classification accuracies (\%) on 5000 test episodes of novel categories (with 95\% confidence intervals). ``Sem." denotes whether to leverage semantic knowledge.}
	\vspace{-0.55cm}
	\begin{center}
		\begin{tabular}{l||l|l||cc|cc}
			\hline
			\multirow{2}{*}{Models} & \multirow{2}{*}{Backbone} & \multirow{2}{*}{Sem.} & \multicolumn{2}{|c|}{miniImageNet} & \multicolumn{2}{|c}{tieredImageNet}\\
			& &  & 5Way-1Shot & 5Way-5Shot & 5Way-1Shot & 5Way-5Shot \\
			\hline\hline
			Matching Networks \textcolor{blue}{(NIPS'16)} \cite{vinyals2016matching} & 4Conv  & No & 43.56$\pm$0.84 & 55.31$\pm$0.73 & - & - \\
			MAML \textcolor{blue}{(ICML'17)} \cite{finn2017model} & 4Conv & No & 48.70$\pm$1.84 & 63.11$\pm$0.92 & 51.67$\pm$1.81 & 70.30$\pm$1.75 \\
			ProtoNet \textcolor{blue}{(NIPS'17)} \cite{snell2017prototypical} & 4Conv & No & 49.42$\pm$0.78 & 68.20$\pm$0.66 & 53.31$\pm$0.89 & 72.69$\pm$0.74 \\
			Dynamic-FSL \textcolor{blue}{(CVPR'18)} \cite{gidaris2018dynamic} & 4Conv & No & 56.20$\pm$0.86 & 72.81$\pm$0.62 & - & -  \\
			Dynamic-FSL \textbf{(ours baseline)} & ResNet-12 & No & 62.81$\pm$0.27 & 78.97$\pm$0.18 & 68.55$\pm$0.31 & 83.95$\pm$0.21  \\
			wDAE-GNN \textcolor{blue}{(CVPR'19)} \cite{gidaris2019generating} & WRN-28-10 & No & 61.07$\pm$0.15 & 76.75$\pm$0.11 & 68.18$\pm$0.16 & 83.09$\pm$0.12  \\
			MetaOptNet \textcolor{blue}{(CVPR'19)} \cite{lee2019meta} & ResNet-12 & No & 62.64$\pm$0.61 & 78.63$\pm$0.46 & 65.99$\pm$0.72 & 81.56$\pm$0.53 \\
			DeepEMD \textcolor{blue}{(CVPR'20)} \cite{zhang2020deepemd} & ResNet-12 & No & 65.91$\pm$0.82 & \textbf{82.41}$\pm$0.56 & 71.16$\pm$0.87 & \textbf{86.03}$\pm$0.58 \\
			RFS \textcolor{blue}{(ECCV'20)} \cite{tian2020rethinking} & ResNet-12 & No & 64.82$\pm$0.60 & 82.14$\pm$0.43 & 71.52$\pm$0.69 & \textbf{86.03}$\pm$0.49 \\
			Neg-Cosine \textcolor{blue}{(ECCV'20)} \cite{liu2020negative} & ResNet-12 & No & 63.85$\pm$0.81 & 81.57$\pm$0.56 & - & - \\
			KTN \textcolor{blue}{(ICCV'19)} \cite{peng2019few} & 4Conv & Yes & 64.42$\pm$0.72 & 74.16$\pm$0.56 & -  & -  \\
			TriNet \textcolor{blue}{(TIP'19)} \cite{chen2019multi} & ResNet-18 & Yes & 58.12$\pm$1.37 & 76.92$\pm$0.69 & -  & -  \\
			AM3 \textcolor{blue}{(NIPS'19)} \cite{xing2019adaptive} & ResNet-12 & Yes & 65.30$\pm$0.49 & 78.10$\pm$0.36 & 69.08$\pm$0.47 & 82.58$\pm$0.31  \\
			%AM3 in our framework & ResNet-12 & Yes & 64.29$\pm$0.24 & -$\pm$- & 70.07$\pm$0.30 & -$\pm$-  \\
			%TRAML \textcolor{blue}{(CVPR'20)} \cite{li2020boosting} & ResNet-12 & Yes & 67.10$\pm$0.52
			% & 79.54$\pm$0.60 & - & -  \\
			\hline
			SEGA \textbf{(ours)} & ResNet-12 & Yes & \textbf{69.04}$\pm$0.26 & 79.03$\pm$0.18 & \textbf{72.18}$\pm$0.30 & 84.28$\pm$0.21 \\
			\hline
		\end{tabular}
	\end{center}
	\label{table:mini_result}
	\vspace{-0.65cm}
\end{table*}

\begin{table}[ht]
	\caption{Results on CUB. Test setting is the same as above.}
	\vspace{-0.25cm}
	\label{table:cub}
	\resizebox{1\linewidth}{!}{
		\begin{tabular}{l|cc}
			\hline
			\multirow{2}{*}{Models}  & \multicolumn{2}{c}{CUB}    \\
			& 5Way 1Shot     & 5Way 5Shot     \\ \hline
			TriNet \textcolor{blue}{(TIP'19)} \cite{chen2019multi}      &  69.61$\pm$0.46 & 84.10$\pm$0.35   \\ 
			MultiSem \textcolor{blue}{(CoRR'19)} \cite{schwartz2019baby}      &  76.1$\pm$n/a & 82.9$\pm$n/a   \\ 
			FEAT \textcolor{blue}{(CVPR'20)} \cite{ye2020few}      &  68.87$\pm$0.22 & 82.90$\pm$0.15   \\
			DeepEMD \textcolor{blue}{(CVPR'20)} \cite{zhang2020deepemd}      &  75.65$\pm$0.83 & 88.69$\pm$0.50   \\ \hline
			SEGA \textbf{(ours)}  & \textbf{84.57}$\pm$0.22 & \textbf{90.85}$\pm$0.16 \\ \hline
		\end{tabular}
	}
	\vspace{-0.25cm}
\end{table}

\begin{table}[ht]
	%\caption{Comparisons with popular FSL approaches in average classification accuracies (\%) on CIFAR-FS. The test setting is in the same manner as above.}
	\caption{CIFAR-FS results.Test setting is the same as above.}
	\vspace{-0.25cm}
	\label{table:cifarfs}
	\resizebox{1\linewidth}{!}{
		\begin{tabular}{l|cc}
			\hline
			\multirow{2}{*}{Models}  & \multicolumn{2}{c}{CIFAR-FS}    \\
			& 5Way 1Shot     & 5Way 5Shot     \\ \hline
			MAML \textcolor{blue}{(ICML'17)} \cite{finn2017model}            & 58.9$\pm$1.9   & 71.5$\pm$1.0   \\
			ProtoNet \textcolor{blue}{(NIPS'17)} \cite{snell2017prototypical}              & 55.5$\pm$0.7   & 72.0$\pm$0.6   \\
			%Dynamic-FSL \textbf{(ours baseline)} & 72.06$\pm$0.30   & 85.76$\pm$0.20   \\
			MetaOptNet \textcolor{blue}{(CVPR'19)} \cite{lee2019meta}    & 72.0$\pm$0.7   & 84.2$\pm$0.5   \\
			RFS \textcolor{blue}{(ECCV'20)} \cite{tian2020rethinking}       & 73.9$\pm$0.8   & \textbf{86.9}$\pm$0.5   \\ \hline
			SEGA \textbf{(ours)}     & \textbf{78.45}$\pm$0.24 & 86.00$\pm$0.20 \\ \hline
		\end{tabular}
	}
	\vspace{-0.45cm}
\end{table}

\begin{figure}[t!]
    \vspace{0.15cm}
	\begin{center}
		\includegraphics[width=0.90\linewidth]{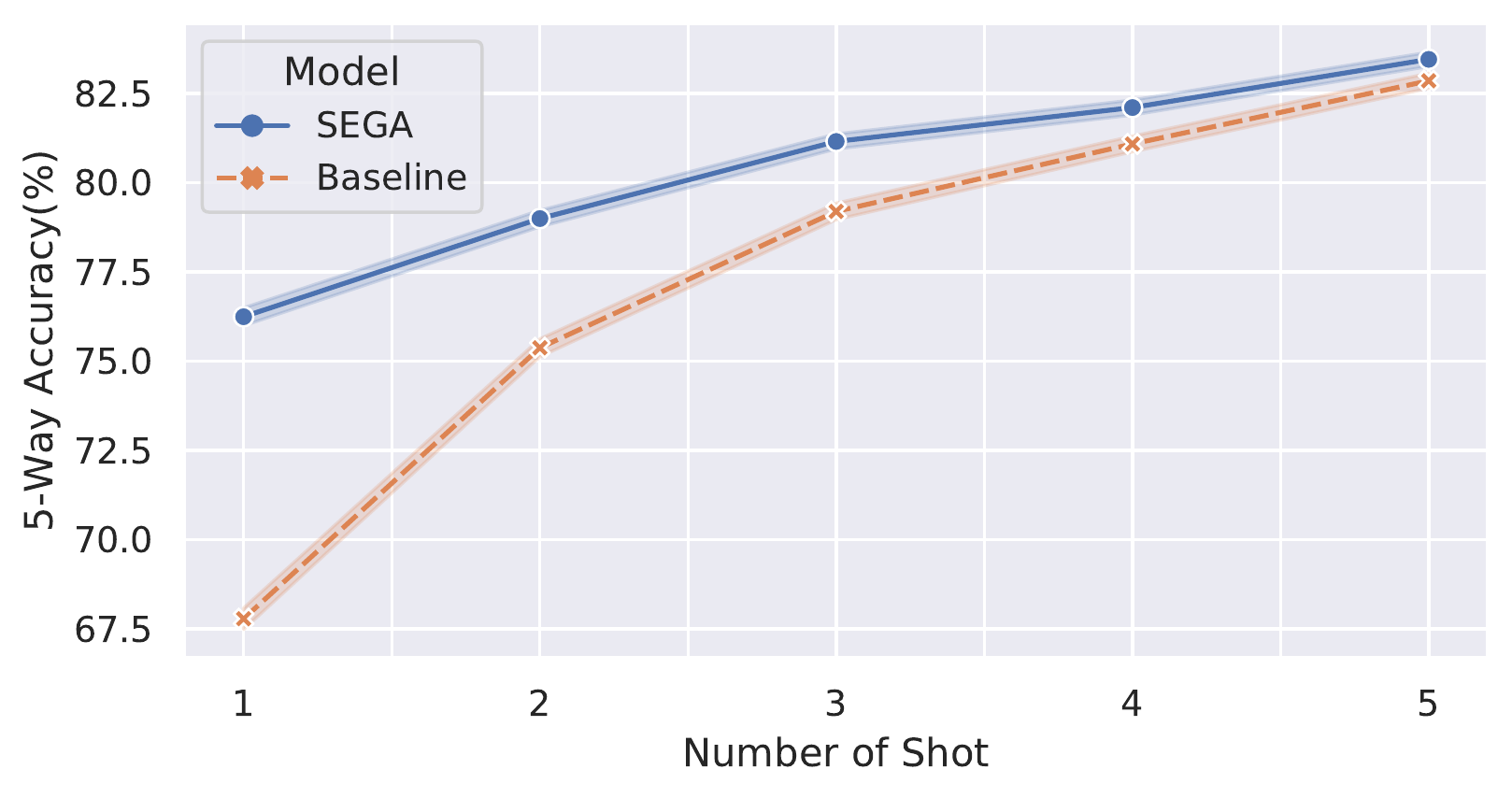}
	\end{center}
	\vspace{-0.75cm}
	\caption{5-Way accuracy on CIFAR-FS from 1 to 5 Shot.}
	\label{fig:1shot-5shot}
	\vspace{-0.55cm}
\end{figure}

\subsection{Benchmark Comparisons and Evaluations}
In this subsection, We make comparison with several popular FSL approaches within the inductive learning framework. Table \ref{table:mini_result} shows the results on miniImageNet and tieredImageNet dataset. Note that KTN, TriNet, AM3, and our SEGA utilize semantic knowledge while other methods are in unimodal settings. Our method achieves the highest performance especially in 5-Way 1-Shot setting and outperforms the most relevant semantic using method AM3 \cite{xing2019adaptive}. Furthermore, we even adapt AM3 to our framework for further fair comparison which can be found in supplementary material. The advantage should be attributed to the more human-like way to utilize knowledge which is the attention mechanism instead of reconstruction of prototype. Figure \ref{fig:1shot-5shot} shows the performance gain from our SEGA is getting down when the number of shots goes larger. The reason is that when only given one sample per class the visual prototype is poor and unstable thus the semantic knowledge can help a lot. However, in 5-Shot setting, the visual prototype is getting stable and accurate when given more samples and the gain from semantic information is getting lower (we show the visualization result of 5-Shot in supplementary material which implies that the generated prototypes are already very stable in visual space when given 5 samples). Even though, our performance can still have an advantage over SOTAs in larger shot scenarios when using purer semantic knowledge (\eg CUB attributes) to guide the attention (as the CUB results shown in Table \ref{table:cub}). Results on CIFAR-FS are shown in Table \ref{table:cifarfs}, where our method also gets competitive results. We also show our advantage over other SOTAs in computation complexity in the supplementary material.

%------------------------------------------------------------------------
\section{Conclusions}
In this work, we propose a simple yet effective FSL approach which is accomplished by \textbf{SEmantic Guided Attention (SEGA)} on the visual prototype to give top-down guidance on which key features we should focus on. Our proposed approach shows its effectiveness in four popular FSL benchmarks especially when given only one labeled sample for each novel category. Furthermore, we dive deep into how and why our semantic attention works, and further conduct extensive and interesting experiments. Besides, we also analyze the correlation of visual space and semantic space and find out that the alignment calculated on base classes can be transferred and generalized well to novel classes which gives fundamental evidence for the usefulness of the semantic knowledge for FSL task.

\vspace{0.1cm}
\noindent\textbf{Acknowledgements.} \small{This work is partially supported by Natural Science Foundation of China under contracts Nos. U19B2036, 61922080, 61772500, CAS Frontier Science Key Research Project No. QYZDJ-SSWJSC009, and National Key R\&D Program of China (2020AAA0105200).}

{\small
\bibliographystyle{ieee_fullname}
\bibliography{egbib}
}

\end{document}